# Content Based Image Retrieval System using Feature Classification with Modified KNN Algorithm

T. Dharani [#1] IEEE Member, I. Laurence Aroquiaraj [*2] IEEE Member

*Assistant Professor*
[*] *Department of Computer Science, Periyar University*
*Salem - 636011, Tamil Nadu, India*

[#] *Department of Computer Science, Periyar University*
*Salem - 636011, Tamil Nadu, India*

[1] `dharanimca28@gmail.com`
[2] `laurence.raj@gmail.com`

*Abstract*— Feature means countenance, remote sensing scene objects with similar characteristics, associated to interesting scene elements in the image formation process. They are classified into three types in image processing, that is low, middle and high. *Low level* features are color, texture and *middle level* feature is shape and *high level* feature is semantic gap of objects. An image retrieval system is a computer system for browsing, searching and retrieving images from a large image database. Content Based Image Retrieval (CBIR) is a technique which uses visual features of image such as color, shape, texture, etc…to search user required image from large image database according to user's requests in the form of a query. MKNN is an enhancing method of KNN. The proposed KNN classification is called MKNN. MKNN contains two parts for processing, they are validity of the train samples and applying weighted KNN. The validity of each point is computed according to its neighbors. In our proposal, Modified K-Nearest Neighbor (MKNN) can be considered a kind of weighted KNN so that the query label is approximated by weighting the neighbors of the query. The procedure computes the fraction of the same labeled neighbors to the total number of neighbors. MKNN classification is based on validated neighbors who have more information in comparison with simple class labels. This paper also concentrates identifying the unlabeled images with help of MKNN algorithm. Experiments show the validity takes into accounts the value of stability and robustness of the any train samples regarding with its neighbors and excellent improvement in the performance of KNN method. This system allows provide label to unlabeled image as user input.

## I. INTRODUCTION

An image retrieval system is a computer system for browsing, searching and retrieving images from a large database of digital images. Most traditional and common methods of image retrieval utilize some method of adding metadata such as captioning, keywords, or descriptions to the images so that retrieval can be performed over the annotation words. Applications of image retrieval are remote sensing, fashion, crime prevention, publishing, medicine, architecture, etc [1]. It is classifying two types of retrieval are Text Based Image Retrieval and Content Based Image Retrieval.

Initially it working based on text of image. Text Based Image Retrieval is having demerits of efficiency, lose of information, more expensive task and time consuming. Overcome these problems by using Content Based Image Retrieval (CBIR) system for image retrieval. CBIR working by using features of the content of the image is known as Content-Based Image Retrieval (CBIR) [2]. The image retrieval system acts as a classifier to divide the images in the image database into two classes, either relevant or irrelevant. In this sense, an annotated image can be represented by a feature vector x, e.g. a set of image features or eigen features, and its label y that is either relevant or irrelevant [3] and without label image is called unannotated or unlabelled image. These images are not considered during retrieval process. Unlabelled image is most useful to getting the better result of retrieval.

Unlabelled images are identified when classifying the image database by using classifiers. The classifiers are working based on various classification properties of image database. The begin stage of image database classification is supervised. The supervised classification is finding the labeled images very well. The unsupervised classification is used to finding the unlabelled images very well. The section II of this paper presenting basics of Content Based Image Retrieval system. Section III of this paper is tried to explain various existing image classification methods with unlabelled image. Section IV is presenting proposed method of image classification of CBIR system. Section V is including the conclusion of the paper.

## II. CONTENT BASED IMAGE RETRIEVAL

Content based image retrieval is working with different types of image database. All databases are having two types of images like labelled and unlabelled.

A. Example of Labelled and Unlabelled images

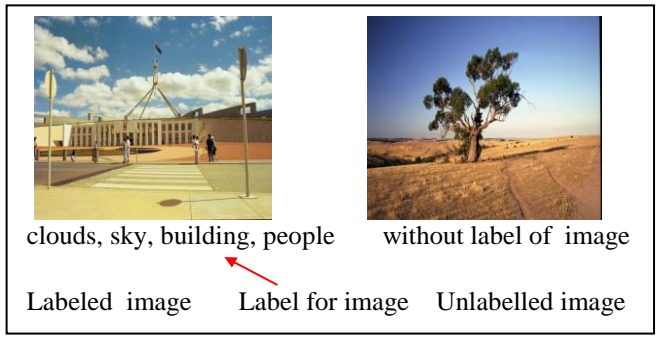

clouds, sky, building, people    without label of image

Labeled image    Label for image    Unlabelled image

B. Basic Architecture of CBIR

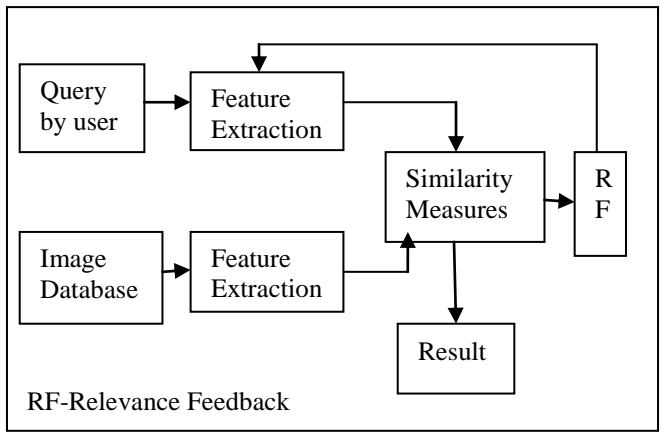

C. Image Database

Content Based Image Retrieval systems are working with various type of image file formats[4] and also different types of size, themes, features, information, etc. The image file formats are

- TIFF 6.0 (Tagged Image File Format)
- GIF 89a (Graphics Interchange Format)
- JPEG (Joint Photographic Expert Group)
- JFIF (JPEG File Interchange Format)
- JP2-JPX
- JPEG 2000
- Flashpix 1.0.2
- ImagePac, Photo CD
- PNG 1.2 (Portable Network Graphics)
- PDF 1.4 (Portable Document Format)

D. Query

An important parameter to measure user-system interaction level[4] is the complexity of queries supported by the system. **From a user perspective,**

*Keywords.* This is a search in which the user poses a simple query in the form of a word or bigram.
*Free-Text.* This is where the user frames a complex phrase, sentence, question, or story about what she desires from the system.
*Image.* Here, the user wishes to search for an image similar to a query image.
*Graphics.* This consists of a hand-drawn or computer-generated picture, or graphics could be presented as query.
*Composite.* These are methods that involve using one or more of the aforesaid modalities for querying a system. This also covers interactive querying such as in relevance feedback systems.
The processing becomes more complex when visual queries and/or user interactions[4] are involved. We next broadly characterize query processing **from a system perspective,**

*Text-Based.* Text-based query processing usually boils down to performing one or more simple keyword-based searches and then retrieving matching pictures.
*Content-Based.* Content-based query processing lies at the heart of all CBIR systems. Processing of query (image or graphics) involves extraction of visual features and/or segmentation and search in the visual feature space for similar images.
*Composite.* Composite processing may involve both content- and text-based processing in varying proportions.
*Interactive-Simple.* User interaction using a single modality needs to be supported by a system.
*Interactive-Composite.* The user may interact using more than one modality (e.g., text and images).

E. Feature Extraction

Feature means countenance, characteristics of object. Feature extraction is refers that dimensionality reduction of that object. In general it defines when performing analysis of complex data one of the major problems stems from the number of variables involved. It plays an important role in image processing. Features are classified into three types in image processing, that is low, middle and high. *Low level* features are color, texture and *middle level* feature is shape and *high level* feature is semantic gap of objects. Feature vector is used to store the extracted features as matrix format in database for processing. Feature extraction is working with matrix, histogram, Transformation, Rotation, Segmentation, Region of Interest, Region of Center of the image.

F. Color Feature

CBIR system is working with mainly one common feature of image is color. The color images are having the standard color is RGB color. Here using the coloring model of image database. First of all separating the original image as RGB color form by using MATLAB code for best retrieval process.

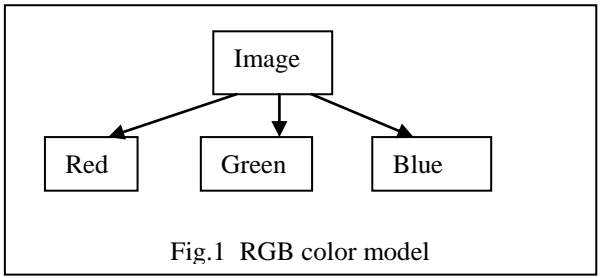

Fig.1 RGB color model

```
I1=imread('Image01.jpg');
Red color :   r = I1(:,:,1);
Green color:  g = I1(:,:,2);
Blue color:   b = I1(:,:,3);
```

Fig.2 MATLAB code for RGB color model

### G. Similarity Measures

CBIR systems are working with some measures for calculating distance between similarities of two images during image retrieval time. The measures are sometimes similar or otherwise dissimilar between two images. CBIR system frequently used similarity measures for an accuracy of image retrieval from large image database. CBIR has overviewed with various similarities measures. These distance measures are mainly used for identify, sum, calculate of similarities distance between images, regions, features vectors, color histograms etc. Content Based Image Retrieval includes the standard measures. Those measures are the initial measures of the image retrieval process. The measures are *Recall, Precision, and Fallout*. These measures are used to improve the image retrieval process with various measures.

**Table I**
**Standard information retrieval measures**

| Measures | Equations | |
|---|---|---|
| Recall | $Recall = \frac{Total\ number\ of\ retrieved\ relevant\ images}{Total\ number\ of\ relevant\ images}$ | (1) |
| Precision | $Precision = \frac{Total\ number\ of\ retrieved\ relevant\ images}{Total\ number\ of\ retrieved\ images}$ | (2) |
| Fallout | $F = \frac{False\ alarms}{False\ alarms + Correct\ dismissals}$ | (3) |

The similarity measures include a benchmark measures that is Euclidean distance between two similar images. This measure is used to improve the image retrieval process in an efficient way with various types of measures with different combinations of two object of an image. Similarity measures are very useful to calculate the different object of an image for accuracy and fast image retrieving.

### H. Relevance Feedback

Relevance feedback (RF) is a query modification technique which attempts to capture the user's precise needs through iterative feedback[4] and query refinement. It can be thought of as an alternative search paradigm, complementing other paradigms such as keyword-based search.

*Learning-Based Advancements.* Based on the user's relevant feedback, learning-based approaches are used to modify the feature set or similarity measure.
*Feedback Specification Advancements.* Traditionally, RF has engaged the user in multiple rounds of feedback, each round consisting of one set each of positive and negative examples in relation to the intended query.
*User-Driven Methods.* While many past attempts at RF have focused on the machine's ability to learn from user feedback, the user's point-of-view in providing the feedback has largely been taken for granted.
*Probabilistic Methods. Used* where uncertainty about the user's goal is represented by a distribution over the potential goals, following which the Bayes' rule helps in selecting the target image.

## III. EXISTING METHODS

The classification is used to group with homogeneous characteristics, with the multiple objects from each other within the image. Classification divides the feature space into several classes based on a decision rule. *Supervised classification* is the sampling[5] of training data from clearly identified training areas. *Unsupervised classification* is working with two types they are multiple groups from randomly selected sampling data and clustering classes.

### A. Learning

Training examples are *labeled*, this means that the ground-truth labels of them are known to the learner. The usage of unlabeled data to help supervised learning efficiently. The paradigms are used to learning by using unlabeled data in an efficient way.

### B. Semi-supervised learning

Semi-supervised learning deals with methods for automatically exploiting unlabeled data in addition to labeled data to improve learning performance. That is, the exploitation of unlabeled data does not need human intervene. Here the key is to use the unlabeled data to help estimate the data distribution. *Transductive learning[15]* is a cousin of semi-supervised learning, which also tries to exploit unlabeled data automatically.

### C. The Graph Mincut Learning Algorithm

This algorithm is[6] working with weighted graph G = (V, E) by using edge. The classification of vertices is connected by edges of infinite weight to the labelled examples. The weights are assigned based on relationships between example vertices. The minimum cut for weights of the graph the minimum total weights of the edges whose removal disconnects from (v+, v-). Finally assign the positive and negative labels to all unlabelled examples of (v+, v-) [51].

### D. Anchoring Edge Fragments to Local Patches

Unlabeled image as a set of semi-local[7] region features, $X = \{f_1,...,f_{|x|}\}$ where each $f_i$ consists of a local appearance descriptor and all the surrounding edge fragments and their weights.

### E. k-Means

K-means clustering is an algorithm to classify or to group your objects[8] based on attributes/features into K number of group. K is positive integer number. The grouping is done by minimizing the sum of squares of distances between data and the corresponding cluster centroid. Thus the purpose of K-mean clustering is to classify the data.

### F. K Nearest Neighbors Algorithm

K-nearest neighbor is a supervised learning algorithm where the result of new instance query is classified based on majority[9] of K-nearest neighbor category. The purpose of this algorithm is to classify a new object based on attributes and training samples. The classifiers do not use any model to fit and only based on memory. Given a query point, we find K number of objects or (training points) closest to the query point. The classification is using majority vote among the classification of the K objects. Any ties can be broken at random. K Nearest neighbor algorithm used neighborhood classification as the prediction value of the new query instance.

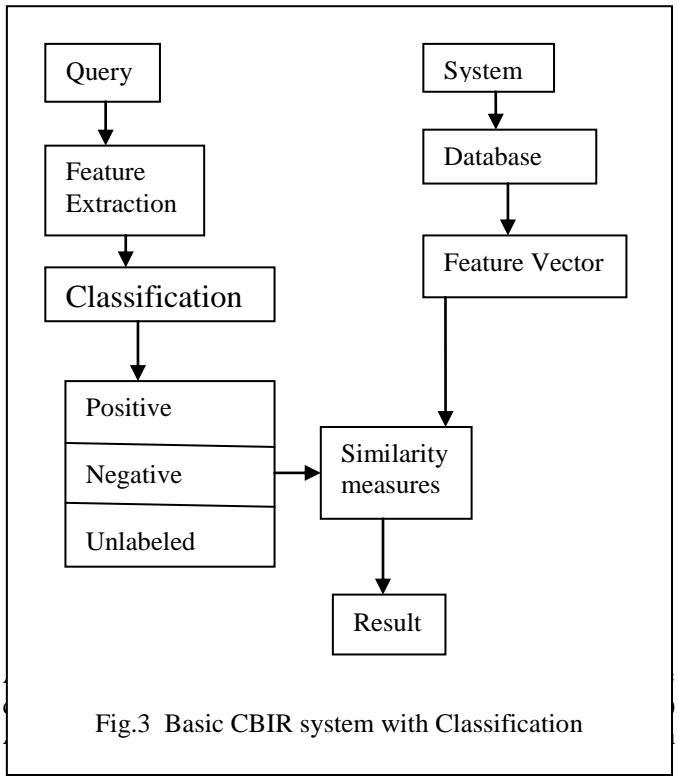

Fig.3 Basic CBIR system with Classification

(EM) with unlabeled image as data set. Support Vector Machines using the approach structural risk minimization with unlabeled image as two classes relevant and irrelevant. Relevance Feedback using the approach high confidence degree with unlabeled image as rank. BMMA AND SEMIBMMA FOR SVM RF[11] using the approach Laplacian regularizer with unlabeled image as sample. Semi-Supervised/Active learning using the approach co-training paradigm, co-testing paradigm with unlabeled image as training set. SSAIR (Semi-Supervised Active Image Retrieval) using the approach Shrunk with unlabeled image as data set. TRANSDUCTIVE LEARNING METHOD[12] using the approach Probability with unlabeled image as ranking scores. ACTIVE LEARNING METHOD using the approach Convergence with unlabeled image as two examples (positive for relevant and negative for irrelevant). Bootstrapping SVM active learning using the approach images are ranked in descending order with unlabeled image as initial SVM classifier. Local Weakly Similar Patches using the approach Semi-supervised learning algorithm with unlabeled image as labeled log image. SEMANTIC KERNEL LEARNING[10] using the approach Factorization with unlabeled image as kernel matrix. Semi-Supervised Multiple Instance Learning (SSMIL)[13] using the approach Optimization with unlabeled image as instance of mapped bag. New Analysis of the Value of Unlabeled Data using the approach Distribution with unlabeled image as positive and negative examples. Label propagation using the approach Binary approach with unlabeled image as fuzzy relevance. The Graph Mincut Learning Algorithm using the approach Weighted graph with unlabelled data as example vertices. Self-taught Learning Algorithm using the approach Correlations between rows of pixels with unlabelled data as example data. Anchoring Edge Fragments to Local Patches using the approach SIFT descriptor with unlabeled image as a set of semi-local region features.

## IV. PROPOSED METHOD

In proposed system using Modified K-Nearest Neighbor Algorithm for image classification in Content Based Image Retrieval system. Algorithm consists of two parts: **VALIDITY OF THE TRAIN SAMPLES** that takes every training sample must be validated at the first step. The validity of each point is computed according to its neighbors. **APPLYING WEIGHTED KNN** means weighted KNN is one of the variations of KNN method which uses the K nearest neighbors, regardless of their classes, but then uses weighted votes from each sample rather than a simple majority or plurality voting rule.

### A. Modified K-Nearest Neighbor Algorithm

Inspired the traditional KNN algorithm, the main idea is classifying the test[16] samples according to their neighbor tags. This method is a kind of weighted KNN so that these weights are determined using a different procedure. The procedure computes the fraction of the same labeled neighbors to the total number of neighbors.

The main idea of the presented method is assigning the class label of the data according to *K* validated data points of the train set. In other hand, first, the validity of all data samples in the train set is computed.

```
Output_label := MKNN ( train_set , test_sample )
Begin
For i := 1 to train_size
Validity(i) := Compute Validity of i-th sample;
End for;
Output_label:=Weighted_KNN(Validity,test_sample);
Return Output_label ;
End.
```

Fig. 4. Pseudo-code of the MKNN Algorithm

### B. *Validity of the Train Samples*

The validation process is performed for all train samples once. After assigning the validity of each train sample, it is used as more information about the points. To validate a sample point in the train set, the *H* nearest neighbors of the point is considered. Among the *H* nearest neighbors of a train sample *x*, validity(*x*) counts the number of points with the same label to the label of *x*.

$$Validity(x) = \frac{1}{H} \sum_{i=1}^{H} S(lbl(x), lbl(N_i(x))) \quad (1)$$

where *H* is the number of considered neighbors and *lbl(x)* returns the true class label of the sample *x*. also, *Ni(x)* stands for the *i*th nearest neighbor of the point *x*. The function *S* takes into account the similarity between the point *x* and the *i*th nearest neighbor.

### C. *Applying Weighted KNN*

Each of the *K* samples is given a weighted vote that is usually equal to some decreasing function of its distance from the unknown sample. For example, the vote might set be equal to *1/(de+1)*, where *de* is Euclidian distance. These weighted votes are then summed for each class, and the class with the largest total vote is chosen. This distance weighted KNN technique is very similar to the window technique for estimating density functions. For example, using a weighted of *1/(de+1)* is equivalent to the window technique with a window function of *1/(de+1)* if *K* is chosen equal to the total number of training samples. In the MKNN method, first the weight of each neighbor is computed using the *1/(de+0.5)*.

Then, the validity of that training sample is multiplied on its raw weight which is based on the Euclidian distance. In the MKNN method, the weight of each neighbor sample is derived according to

$$W(i) = Validity(i) \times \frac{1}{d_e + 0.5} \quad (2)$$

where *W(i)* and *Validity(i)* stand for the weight and the validity of the *i*th nearest sample in the train set.

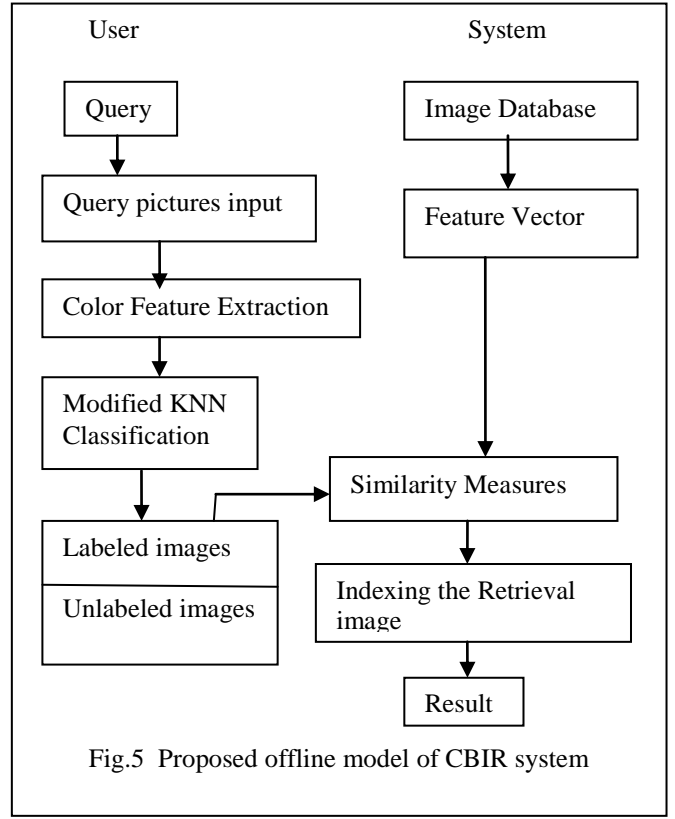

Fig.5 Proposed offline model of CBIR system

## V. CONCLUSION

In this paper we presented CBIR system using feature classification with Modified K-Nearest Neighbor Algorithm. CBIR system based on visual feature of image that color of image enables to classify the image database as labeled and unlabeled. The offline model contains two major parts, namely Validity of the Train Samples, Applying Weighted KNN. MKNN algorithm adds a new value named "Validity" to train samples which cause to more information about the situation of training data samples in the feature space. The validity takes into accounts the value of stability and robustness of the any train samples regarding with its neighbors. Identifying unlabeled image and providing label with help of MKNN classification by using the user input.

ACKNOWLEDGMENT

The second Author of this paper acknowledges UGC-Minor Research Project (No.F.41-1361/2012 (SR)) for the financial support.